# SISTEM PENUNJANG KEPUTUSAN KELAYAKAN PEMBERIAN PINJAMAN DENGAN METODE FUZZY TSUKAMOTO

**Tri Murti[1), ] Leon Andretti Abdillah[2)], Muhammad Sobri[3)]**

[1,2)] Program Studi Sistem Informasi – Fakultas Ilmu Komputer, Universitas Bina Darma
[3)] Program Studi Sistem Informasi DIII – Fakultas Ilmu Komputer, Universitas Bina Darma
Jl. Ahmad Yani No.12, Plaju, Palembang
email:[1)]trimurti444@ymail.com, *[2)]leon.abdillah@yahoo.com, [3)]sobri.irbos@gmail.com

***Abstrak*** – Sistem penunjang keputusan (SPK) dapat digunakan untuk membantu penyelesaikan permasalahan atau pengambilan keputusan yang bersifat semi terstruktur atau terstruktur. Metode yang digunakan adalah Fuzzy Tsukamoto. PT Triprima Finance merupakan suatu perusahaan yang bergerak di bidang jasa peminjaman dengan jaminan berupa Buku Pemilik Kenderaan Bermotor atau mobil (BPKB). PT. Triprima Finance harus mempertimbangkan pinjaman dari para nasabahnya dengan persetujuan dari kepala manajer. Persetujuan tersebut memerlukan waktu yang lama karena harus melewati banyak tahap prosedur laporan. Kegiatan pengambil keputusan pada PT Triprima Finance dilakukan dengan proses analisis secara manual. Untuk membantu mengatasi masalah tersebut maka diperlukannya metode penyelesaian dalam ketepatan dan kecepatan pengambilan keputusan kelayakan pemberian pinjaman. Untuk mengatasi hal tersebut perlu dikembangkan sistem yang baru yaitu sistem pendukung keputusan dengan metode fuzzy tsukamoto. diharapkan dapat mempermudah kaposko untuk menentukan keputusan yang akan diambil.

***Kata kunci*** : Sistem penunjang keputusan (SPK), Fuzzy tsukamoto, Pemberian pinjaman.

## I. PENDAHULUAN

Teknologi informasi (TI) telah diadopsi dalam berbagai bidang kehidupan. Hal ini dimungkinkan karena teknologi komputer mampu berkolaborasi dengan banyak bidang ilmu lainnya [1]. TI telah membawa perubahan yang sangat mendasar bagi organisasi baik swasta maupun publik [2]. Sehingga TI sudah menjadi backbone utama bagi banyak aspek di kehidupan kita sekarang [3]. Salah satunya adalah sistem pendukung keputusan (SPK).

Sistem Pendukung Keputusan (SPK) atau *Computer Based Decision Support System* (DSS) merupakan salah satu bagian dari sistem informasi yang berguna untuk meningkatkan efektifitas pengambilan keputusan. Permasalahan yang umum dijadikan objek pada SPK ada yang bersifat yang bersifat semi terstruktur atau terstruktur. Pada artikel kali ini para penulis akan membahas mengenai pengambilan keputusan di sektor keuangan.

PT Triprima Finance merupakan perusahaan yang bergerak dibidang jasa peminjaman dengan jaminan berupa Buku Pemilik KenderaanBermotor atau mobil (BPKB). Dengan tujuan untuk memenuhi pelayanan yang baik kepada nasabah, PT Triprima Finance harus mempertimbangkan pinjaman dari para nasabahnya dengan persetujuan dari kepala manajer. Persetujuan tersebut dapat memerlukan waktu yang lama karena harus melewati banyak tahap prosedur laporan. Selama ini kegiatan pengambilan keputusan pada PTTriprima *Finance* dilakukan dengan proses analisis secara manual dengan cara mempertimbangkan berdasarkan data nasabah. Untuk membantu mengatasi masalah tersebut maka diperlukanlah metode penyelesaian dalam ketepatan dan kecepatan pengambilan keputusan kelayakan pemberian pinjaman. Dalam sistem penilaian kelayakan yang akan dibangun ini harus ada kriteria-kriteria-nya. Kriteria penilaian kelayakan pemberian pinjaman uang pada PT Triprima *Finance* meliputi penilaian jumlah penghasilan, jumlah pinjaman, dan jaminan. Untuk memudahkan proses tersebut maka penulis menggunakan metode *fuzzy tsukamoto.*

Sejumlah penelitian telah dilakukan yang berhubungan dengan sistem pendukung keputusan antara lain: 1) *Fuzzy Inference System* dengan *tsukamoto* sebagai pemberian saran Pemilihan Konsentrasi Jurusan [4], 2) Sistem pendukung keputusan kelayakan TKI ke luar negeri [5], 3)Sistem Pendukung Keputusan Untuk Menentukan Kelayakan Pemberian Pembiayaan Nasabah [6], 4)Pemberian Beasiswa Bidik Misi menggunakan *Simple Additive Weighting (SAW)* [7], dan 5) Pengambilan keputusan pemberian kredit pemilikan rumah [8].

Metode ini dilakukan untuk membantu manajemen keputusan terhadap pilihan alternatif untuk mendapatkan keputusan yang akurat dan optimal serta dapat memecahkan suatu masalah. Ada beberapa jenis metode yang digunakan dalam pengambilan keputusan contohnya *fuzzy* tahani, *fuzzy* FMADM, dan *fuzzy* SAW. Dalam penelitian ini penulis menggunnkan metode *fuzzy tsukamoto*. Dipilihnya metode *fuzzy tsukamoto* karena didalam metode *fuzzy* adanya derajat keanggotaan yang memiliki rentang nilai 0 hingga 1. Sedangkan *tsukamoto* yang mempunyai aturan berbentuk *IF-THEN* yang akan dipresentasikan dalam himpunan *fuzzy*. Sebagai hasil *output* diinferensikan dari tiap-tiap aturan diberikan dengan berdasarkan predikat. Metode ini diharapkan akan lebih memudahkan dalam pemberian penilaian yang lebih tepat, lebih efektif, mudah dan proses penilaian kelayakan pemberian pinjaman uang.

## II. METODE PENELITIAN

Metode pengembangan perangkat lunak yang digunakan dalam merancang dan membangun perangkat lunak ini adalah *System Development Life Cycle* (SDLC) [9] yang terdiri atas: 1) Investigasi sistem (*System investigation*), 2) Analisis sistem (*System analysis*), 3) Rancangan sistem (*System design*), 4) Implementasi sistem (*System implementation*), dan 5) Perawatan sistem (*System maintenance*).

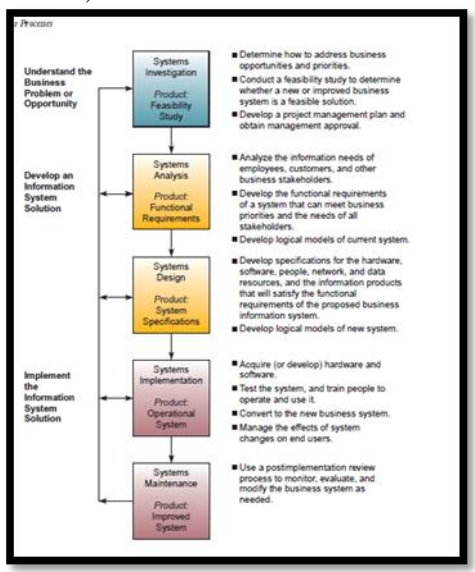

Gambar 1. Tahapan-tahapan dalam *SDLC*

### 2.1 Fuzzy

*Fuzzy* secara bahasa diartikan sebagai kabur atau samar-samar. Dalam *fuzzy* dikenal derajat keanggotaan yang memiliki rentang nilai 0 hingga 1. Berbeda dengan himpunan yang memiliki nilai 1 atau 0. Sedangkan logika *fuzzy* adalah suatu cara yang tepat untuk memetakan suatu ruang input kedalam suatu ruang *output*, mempunyai nilai kontinyu. *Fuzzy* dinyatakan dalam derajat dari suatu keanggotaan dan derajat dari kebenaran. Oleh sebab itu sesuatu dapat dikatakan sebagian benar dan sebagian salah pada waktu yang sama [10].

Variabel *Fuzzy* merupakan variabel yang hendak dibahas dalam suatu sistem Himpunan *Fuzzy*. Dalam himpunan *fuzzy* terdapat beberapa representasi darifungsi keanggotaan, salah satunya yaitu representasi *linear*. Pada representasi *linear*, pemetaan input ke derajat keanggotaannya digambarkan sebagai suatu garis lurus. Berikut ini adalah gambar representasi *linear* rendah dan naik.

### 2.2 Representasi Linear Rendah

Dibawah ini grafik representasi liner rendah pada keanggotaan himpunan *fuzzy* digambarkan sebagai berikut.

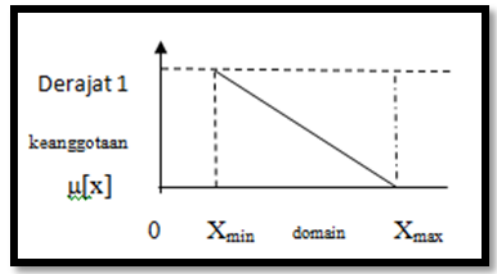

Gambar 2. representasi liner rendah himpunan *fuzzy*

$$\mu rendah\,[x] = \begin{cases} 1 & ; x < X_{min} \\ \frac{Xmax - X}{Xmax - Xmin} & ; X_{min} < x < X_{max} \\ 0 & ; x > X_{max} \end{cases} \quad (1)$$

Konjungsi *fuzzy*
$\mu A \wedge B = \mu A(x) \cap \mu B(y) = \min(\mu A(x), \mu B(y))$ (3)
Disjungsi *fuzzy*
$\mu A \vee B = \mu A(x) \cup \mu B(y) = \max(\mu A(x), \mu B(y))$ (4)

Keterangan :
$\mu A(x)$ ; Derajat keanggotaan dari x dalam A atau derajat x berada dalam A
$\cap$ : Konjungsi
$\cup$ : Diskonjungsi
$\mu B(y)$ : Derajat keanggotaan dari y dalam B atau derajat y berada dalam B

### 2.3 Metode Tsukamoto

Tsukamoto [10] yaitu setiap konsekuen pada aturan berbentuk *IF-THEN* harus dipresentasikan dengan suatu himpunan *fuzzy*, dengan fungsi keanggotaan yang monoton. Sebagai hasilnya, *output* hasil inferensi dari tiap-tiap aturan diberikan dengan berdasarkan predikat (*fire strength*). Hasil akhir diperoleh dengan menggunakan rata-rata terbobot. Misalkan ada 2 variabel *input*, yaitu x dan y serta satu variabel output z. Variabel x terbagi atas dua himpunan yaitu $A_1$ dan $A_2$, sedangkan variabel y terbagi atas himpunan $B_1$ dan $B_2$. Variabel z juga terbagi atas dua himpunan yaitu $C_1$ dan $C_2$. Tentu saja himpunan C1 dan C2 harus merupakan himpunan yang bersifat monoton. Ada 2 aturan yang digunakan, yaitu:

$[R1] IF\,(x\,is\,A_1)\,And\,(y\,is\,B_2)\,THEN\,(z\,is\,C_1)$ (5)
$[R1] IF\,(x\,is\,A_2)\,And\,(y\,is\,B_2)\,THEN\,(z\,is\,C_2)$ (6)

Keterangan:
R1 : Aturan *fuzzy*
x : variabel pinjaman
$\alpha_1$ : Himpunan pinjaman tertinggi
$\alpha_2$ : Himpunan pinjaman terendah
And : Operator yang digunakan

y : Variabel jaminan
$B_1$ : Himpunan jaminan tertinggi
$B_2$ : Himpunan jaminan terendah
THEN : Operator yang digunakan
Z : Variabel Penghasilan (nilai crisp)
$C_1$ : Himpunan penghasilan tertinggi (harus monoton)
$C_2$ : Himpunan penghasilan terendah (harus monoton)

Proses inferensi dengan menggunakan metode Tsukamoto [11] sebagaimana terlihat pada gambar 2.

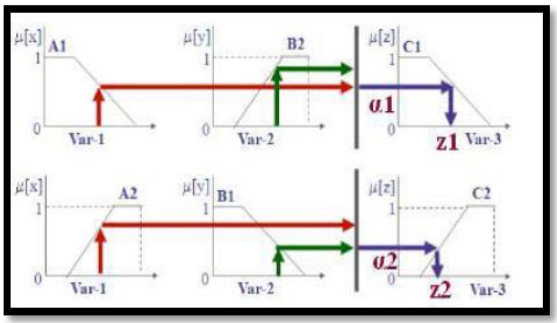

**Gambar 3. Proses *inferensi* dengan menggunakan metode Tsukamoto**

### 2.4 Rancangan *Use Case Diagram*

*Use case* diagram digunakan untuk melihat berhubungan langsung antara pengguna sistem. Yaitu administrasi, *surveyor*, kaposko, dan *branch manager*. Admin dapat melakukan *login*, mengelolah data nasabah, melihat laporan hasil *survey*, setelah itu admin dapat melakukan konfirmasi hasil *survey*, dan simpan hasil *fuzzy tsukamoto* serta dapat melihat hasil penilaian. Bagian *surveyor* dapat mencetak data nasabah dan menginput hasil *survey*. Untuk bagian kaposko dapat melihat data nasabah, melihat laporan hasil *survey* dan dapat melihat hasil penilaian. Sedangkan bagian *branch manager* juga dapat melihat laporan data nasabag, hasil *survey* dan hasil penilaian.

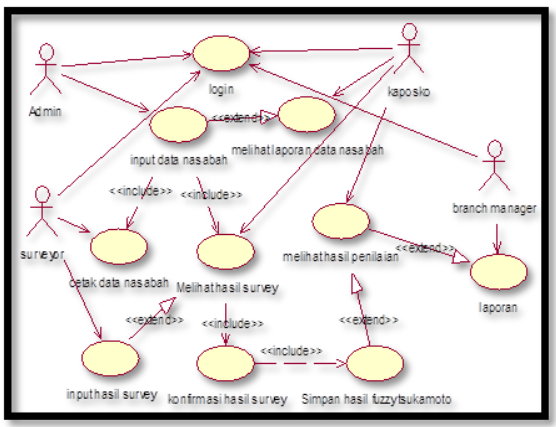

**Gambar 4.*Use Case Diagram***

## III. HASIL DAN PEMBAHASAN

Setelah dilakukan kegiatan-kegiatan yang sesuai dengan metode pengembangan SDLC, maka didapatkan hasil berupa sistem penunjang keputusan pemberian pinjaman.

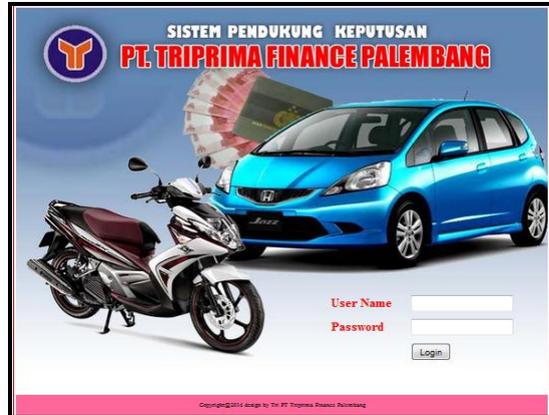

**Gambar 5. Halaman *login***

### 3.1 Input Data Nasabah

Halaman *input* data nasabah ini akan menampilkan *form* untuk menginput data nasabah yang melakukan pinjaman uang di PT Triprima *Finance* Palembang. Dalam menu ini hanya admin yang bisa menginput data lengkap nasabah. Berikut ini adalah tampilan menu *input* data nasabah.

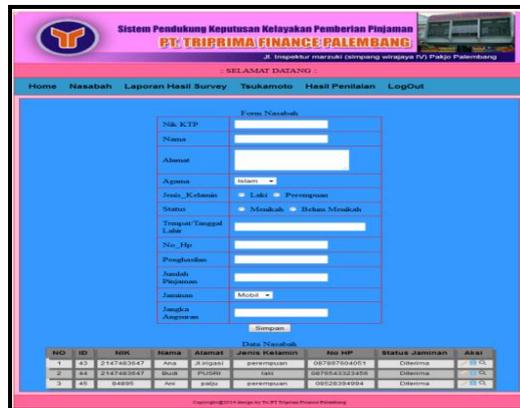

**Gambar 8. Halaman menu *input* data nasabah**

### 3.2 Input Hasil Survey

Untuk menginput hasil *survey*, *surveyor*mengklik menu input hasil *survey* maka akan menampilkan halaman menu input hasil *survey* yang digunakan untuk menginput dan mengelola hasil *survey*. Gambar 9 adalah tampilan halaman hasil *survey*.

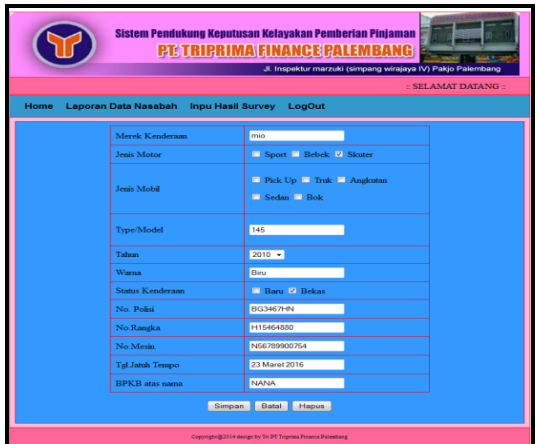

**Gambar 9. Menu input hasil *survey***

### 3.3. Fuzzy Tsukamoto

Untuk melihat *fuzzy tsukamoto*, admin meng*klik* hitungan *fuzzy tsukamoto* maka akan menampilkan halaman menu*fuzzy tsukamoto* yang digunakan hitungan untuk mendapatkan hasil keputusan kelayakan pemberian pinjaman. Berikut ini adalah tampilan halaman metode *fuzzytsukamoto*.

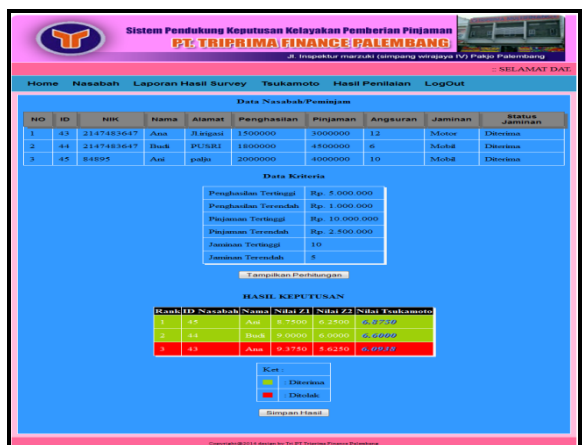

**Gambar 10. Halaman menu *fuzzy tsukamoto***

### 3.4 Hasil Penilaian

Untuk melihat hasil penilaian, *branch manager*mengklik *from* hasil penilaian maka akan menampilkan halaman menu hasil penilaian yang digunakan untuk melihat hasil dari nasabah yang diterima atau ditolak berdasarkan hasil metode *fuzzy tsukamoto*. Berikut ini adalah tampilan halaman hasil penilaian.

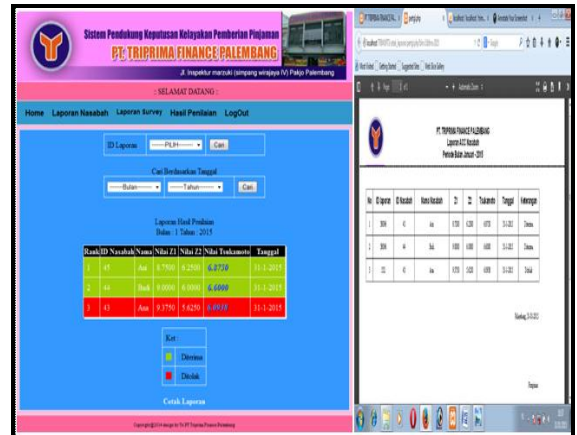

**Gambar 11. Menu hasil penilaian**

## IV. KESIMPULAN

Berdasarkan hasil penelitian yang telah penulis lakukan pada PT Triprima *Finance* Palembang, maka dapat disimpulkan sebagai berikut :

1. Sistem ini dibuat untuk digunakan sebagai sistem pendukung keputusan kelayakan pemberian pinjaman pada PT Triprima *Finance* Palembang dengan metode *fuzzy tsukamoto*. Sehingga mempermudahkan manajamen dalam menentukan kelayakan pinjaman agar lebih cepat dan akurat.
2. Sistem ini juga mempermudahkan dalam memberikan laporan kepada kaposko dan *branch manager*, serta mempermudahkan *surveyor* dalam memberikan laporan hasil *survey*.
3. Sistem ini akan menghasilkan nilai dan keputusan untuk menentukan kelayakan pemberian pinjaman. Berdasarkan nilai total tertinggi dari pengajuan pinjaman.

## DAFTAR REFERENSI

[1] L. A. Abdillah*, et al.*, "Pengaruh kompensasi dan teknologi informasi terhadap kinerja dosen (KIDO) tetap pada Universitas Bina Darma," *Jurnal Ilmiah MATRIK,* vol. 9, pp. 1-20, April 2007.

[2] L. A. Abdillah and D. R. Rahardi, "Optimalisasi pemanfaatan teknologi informasi dalam menumbuhkan minat mahasiswa menggunakan sistem informasi," *Jurnal Ilmiah MATRIK,* vol. 9, pp. 195-204, 2007.

[3] L. A. Abdillah, "Managing information and knowledge sharing cultures in higher educations institutions," in *The 11th International Research Conference on Quality, Innovation, and Knowledge Management (QIK2014)*, The Trans Luxury Hotel, Bandung, Indonesia, 2014.

[4] A. Z. Rakhman*, et al.*, "Fuzzy Inference System dengan Metode Tsukamoto sebagai Pemberi Saran Pemilihan Konsentrasi (Studi Kasus: Jurusan Teknik Informatika UII)," in *Seminar Nasional Aplikasi Teknologi Informasi 2012 (SNATI 2012)*, Yogyakarta, 2012.

[5] A. Ariani, L.A. Abdillah, F. Syakti, "Sistem pendukung keputusan kelayakan TKI ke luar negeri menggunakan FMADM," *Jurnal Sistem Informasi (SISFO),* vol. 4, pp. 336-343, September 2013.

[6] R. Yuniardi, "Perancangan Sistem Pendukung Keputusan Untuk Menentukan Kelayakan Pemberian Pembiayaan Nasabah Baitul Maalwat-Tamwil (BMT) Mujahidin Pontianak Dengan Menggunakan Fuzzy Inference System Metode Tsukamoto," *Jurnal Sistem dan Teknologi Informasi (JustIN),* vol. 2, 2013.

[7] P. Umami*,* L.A. Abdillah, I.Z. Yadi, "Sistem penunjang keputusan pemberian beasiswa bidik misi," in *Konferensi Nasional Sistem Informasi (KNSI)*, STMIK Dipanegara Makassar, Sulawesi Selatan, 2014.

[8] W. Kaswidjanti, "Implementasi Fuzzy Inference System Metode Tsukamoto pada Pengambilan Keputusan Pemeberian Kredit Pemilikan Rumah," *Telematika,* vol. 10, 2014.

[9] J. A. O'Brien, "Introduction To Information Systems". New York: McGraw-Hill/Irwin, 2010.

[10] S. Kusumadewi, "Artificial Intelligence (Teknik dan Aplikasinya)," Yogyakarta: Graha Ilmu*,* 2003.

[11] J.-S. R. Jang, et al., Neuro-Fuzzy and Soft Computing: A Computational APproach to Learning and Machine Intelligence. New Jersey: Prentice-Hall, 1997.

**Biodata Penulis**

*Tri Murti*, memperoleh gelar Sarjana Komputer (S.Kom.), Program Studi Sistem Informasi – Fakultas Ilmu Komputer, Universitas Bina Darma Palembang, lulus tahun 2015.

*Leon Andretti Abdillah*, memperoleh gelar lulusan terbaik Sarjana Komputer, Program Sistem Informasi STMIK Bina Darma Palembang, lulus tahun 2001.Gelar lulusan terbaik program Magister Manajemen Sistem Informasi, Program pascasarjana Universitas Bina Darma Palembang, lulus tahun 2006. Pernah melanjutkan pendidikan PhD di *The University of Adelaide*, Australia pada *School of Computer Science* dengan bidang peminatan *Information Retrieval* (2010-2012). Saat ini menjadi lektor kepala (*Associate Professor*) pada Fakultas Ilmu Komputer Program Studi Sistem Informasi Universitas Bina Darma Palembang.

*Muhammad Sobri*, memperoleh gelar Sarjana komputer (S.Kom.), Program Studi Teknik Informatika Univeristas Bina Darma (UBD), lulus tahun 2009. Tahun 2011 memperoleh gelar Magister Komputer (M.Kom.) dari Konsentrasi Software Engineering UBD. Saat ini sebagai Staf Pengajar program studi Manjemen Informatika UBD Palembang.